\documentclass[10pt, a4paper]{article}
\usepackage{lrec2022} 
\usepackage{multibib}
\newcites{languageresource}{Language Resources}
\usepackage{graphicx}
\usepackage{tabularx}
\usepackage{soul}
\usepackage{expex}
\usepackage{float}
\usepackage{titlesec}
\titleformat{\section}{\normalfont\large\bfseries\center}{\thesection.}{1em}{}
\titleformat{\subsection}{\normalfont\SmallTitleFont\bfseries\raggedright}{\thesubsection.}{1em}{}
\titleformat{\subsubsection}{\normalfont\normalsize\bfseries\raggedright}{\thesubsubsection.}{1em}{}
\renewcommand\thesection{\arabic{section}}
\renewcommand\thesubsection{\thesection.\arabic{subsection}}
\renewcommand\thesubsubsection{\thesubsection.\arabic{subsubsection}}

\usepackage{epstopdf}
\usepackage[utf8]{inputenc}

\usepackage{hyperref}
\usepackage{xstring}

\usepackage{color}

\title{Charon: a FrameNet Annotation Tool for Multimodal Corpora}

\name{Frederico Belcavello\textsuperscript{1}, Marcelo Viridiano\textsuperscript{1}, Ely Edison Matos\textsuperscript{1}, Tiago Timponi Torrent\textsuperscript{1,2}} 

\address{\textsuperscript{1} FrameNet Brasil Lab, Graduate Program in Linguistics, Federal University of Juiz de Fora \\
\textsuperscript{2} Brazilian National Council for Scientific and Technological Development – CNPq \\
         \{fred.belcavello, ely.matos, tiago.torrent\}@ufjf.br, barros.marcelo@estudante.ufjf.br}

\abstract{
This paper presents Charon, a web tool for annotating multimodal corpora with FrameNet categories. Annotation can be made for corpora containing both static images and video sequences paired – or not – with text sequences. The pipeline features, besides the annotation interface, corpus import and pre-processing tools. 
 \\ \newline  }

\begin{document}

\maketitleabstract

\section{Introduction}

Multimodality refers to the property of any communication phenomenon where two or more modes – defined as experientially recognized resources for meaning-making shaped by society and culture – are brought into play \cite{jewitt2003multimodal,kress2010multimodality,bateman2017multimodality} . This paper approaches the expansion of FrameNet annotation into the multimodal domain, as proposed in \newcite{belcavello-etal-2020-frame}, by presenting Charon: a semi-automatic, human-in-the-loop tool for annotating static and dynamic images for semantic frames. Charon was developed to meet the following key requirements: (i) compatibility with existing FrameNet software; (ii) annotation of image with FrameNet categories; (iii) linkage of image and textual annotations.

\section{FrameNet Annotation}

\begin{figure}
    \centering
    \includegraphics[width=\columnwidth]{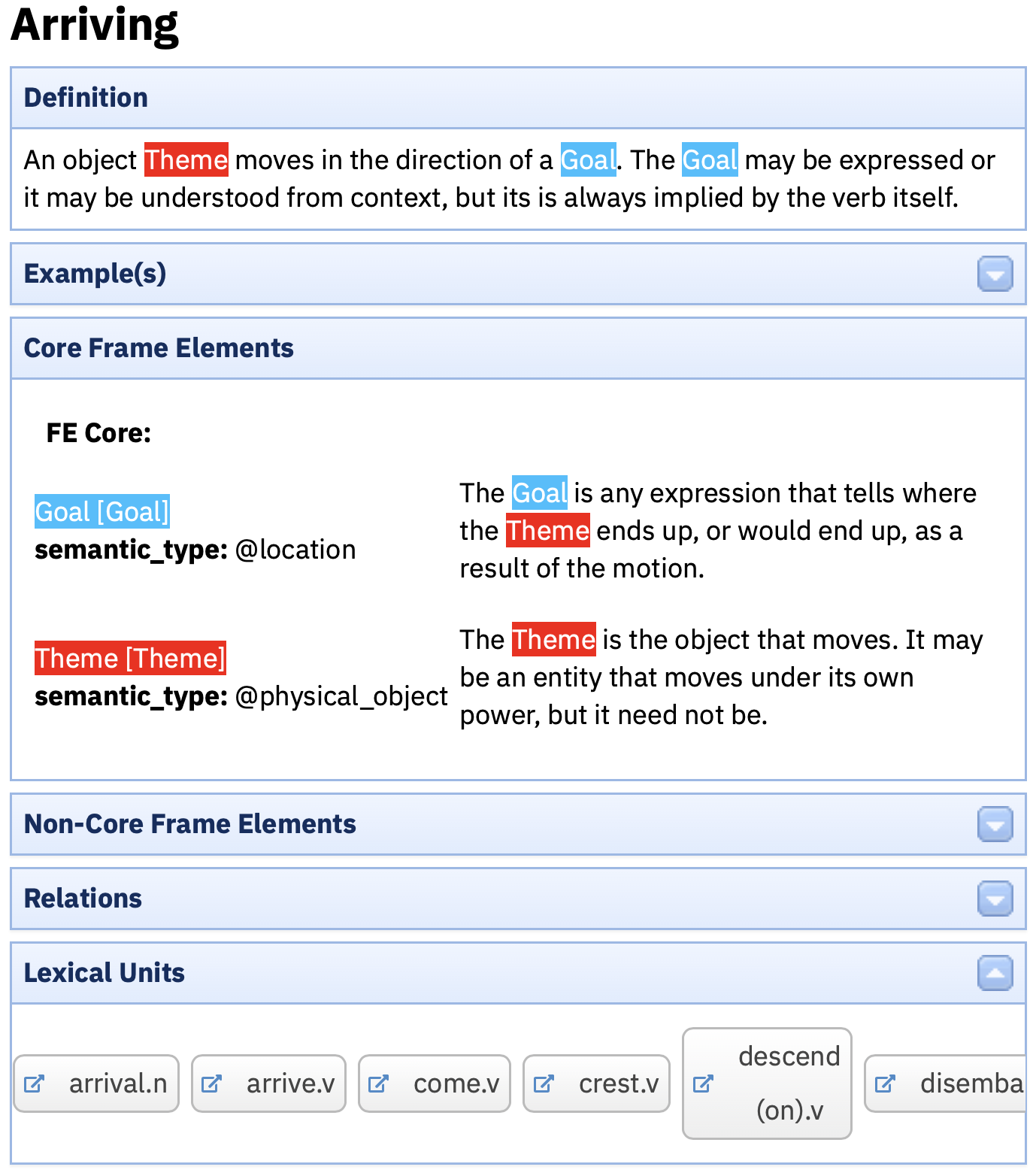}
    \caption{The \texttt{Arrival} frame.}
    \label{fig:arrival}
\end{figure}

FrameNet is a curated language model where lexical items have their meaning defined against systems of concepts called frames \cite{framenet2009}. For instance, words such as \emph{arrive.v} and \emph{arrival.n} have their meanings defined based on a scene where a \textsc{Theme} arrives at a \textsc{Goal}: the \texttt{Arriving} frame (Figure \ref{fig:arrival}). Moreover, frames are connected to one another via a network of typed relations. The \texttt{Arriving} frame, for instance, is inherited by the \texttt{Vehicle\_landing} and preceded by the \texttt{Departing} frames. 

\begin{figure*}[h!]
\centering
\includegraphics[width=\textwidth]{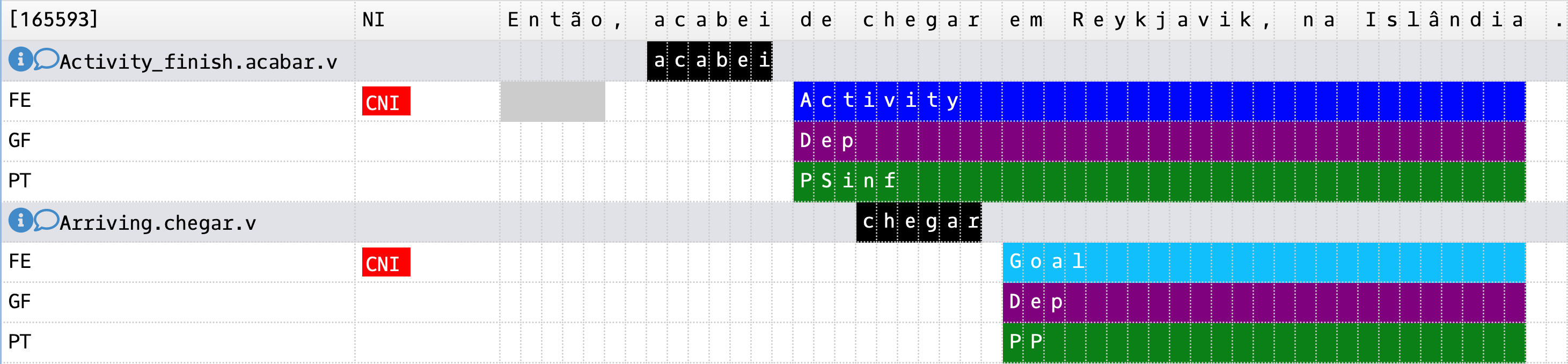}
\caption{\label{fig:sentença}Full-text annotation for sentence (1).}
\end{figure*}

Annotation plays a key role in FrameNet, to the extent that it provides evidence supporting the analysis in the model. Two text annotation methods are used: lexicographic and full-text. In the former, the focus lies on a specific Lexical Unit (LU), and sentences instantiating that LU are extracted from corpora and annotated for a given frame. The aim is to cover the valence patterns of the LU, i.e. its semantic and syntactic affordances. In the latter, the focus is on the corpus being annotated, and the annotator creates Annotation Sets (AS) for each word for which there is an LU in FrameNet. Figure \ref{fig:sentença} shows two of the ASs created for the sentence in (\ref{ex:chegar}).

\ex \label{ex:chegar}Então, acabei de chegar em Reykjavik, na Islândia.\\
\emph{So, I have just arrived in Reykjavik, Iceland.}
\xe

In (\ref{ex:chegar}) the word forms \emph{acabei} and \emph{chegar},  highlighted in black in Figure \ref{fig:sentença}, are the annotation targets. Note that, for each of them, there are three layers of annotation: Frame Element (FE), Grammatical Function (GF) and Phrase Type (PT). The column NI is used for indicating that core FEs are not instantiated in the sentence, but can be inferred.

The idea behind the development of Charon is that other communication modes, namely visual objects, can either evoke frames – similarly to LUs – or complement the valencies of LUs present in text accompanying the images \cite{belcavello-etal-2020-frame}, expanding FrameNet annotation to the multimodal domain. In section \ref{sec:charon}, we describe the tool, but, first, let us turn to a brief summary of other multimodal annotation tools.

\section{Related Work}

The past two decades have witnessed accelerated development of data labeling tools for human annotation of monomodal visual corpora – e.g. COCO Annotator \cite{10.1007/978-3-319-10602-1_48}, ImageTagger \cite{fiedler2018imagetagger}, and LabelBox \cite{sharma2019labelbox}. Moreover, highly generic and flexible multimodal annotation tools, such as Anvil \cite{kipp2001anvil} and ELAN \cite{wittenburg-etal-2006-elan}, allow users to design their own annotation schemes for timeline-based annotation of both audio and visual phenomena from multiple synchronized streams. Finally, frameworks, like SIDGrid \cite{levow2007sidgrid}, extend the functionality of ELAN by allowing the application of user-defined analysis programs to media, time series, and annotations associated with each project. 

Nonetheless, none of these tools and annotation clients allows for the combination of data labeling with the extensive semantic granularity offered by the network of frames and frame elements provided by FrameNet. Allowing for such a combination is the main contribution of Charon, which is presented next.


\section{Charon: Multimodal Annotation Tool} 
\label{sec:charon}

Charon is a multimodal annotation and database management tool. It was developed to annotate visual objects, correlate them with textual data and label frames and Frame Elements evoked by them. Charon is compatible with the FN-Br WebTool: a database management and annotation software used by both local framenet projects in Brazil, Sweden, Croatia and Japan, and in the Global FrameNet Shared Annotation Task \cite{torrent2018multilingual}.\footnote{The FN-Br WebTool is available at \url{https://github.com/FrameNetBrasil/webtool}.} Charon is composed of two modules: a static mode, for annotating picture-text pairings, and a dynamic mode, for annotating video. Both are described next.

\subsection{Annotation of Picture-Caption Pairings}

Charon's static annotation mode can be used to improve multimodal datasets containing picture-text pairings by adding fine-grained semantic information provided by FrameNet. The version of the tool presented in this paper has been tuned to the requirements of the Flickr 30K Entities dataset \cite{7410660} – an expansion of Flickr 30K \cite{young-etal-2014-image}  that adds manually annotated bounding boxes and coreference chains linking entities from each image to their correspondent descriptors in each caption. However, any dataset featuring pictures, captions and bounding boxes identifying parts of the picture can be used.  The annotation process is divided into two stages: (i) corpus import and pre-processing, and (ii)  annotation.  

\subsubsection{Picture Corpus Import and Pre-Processing}

To upload a new corpus, all related files – a folder with JPEG images, a text file with all the sentences, and a XML with the classes and coordinates for each object's bounding box – must be compressed into a ZIP file. Next, Charon creates a new corpus folder in which documents containing lists of image-sentence pairs are built. Before being presented to the annotator, the sentences in these documents are pre-processed by a disambiguation algorithm – DAISY \cite{10.3389/fpsyg.2022.838441} – that identifies and associates each frame-evoking lemma with a semantic frame in the FrameNet database, resulting in an automated frame annotation for each sentence. Such an automated annotation can be checked during a human-in-the-loop process. Charon also checks the image related files for all objects that might have been previously tagged via data labeling tools or computer vision algorithms, and automatically correlates the classes of these objects – obtained from datasets like COCO \cite{10.1007/978-3-319-10602-1_48} and Open Images \cite{OpenImages} – with existing Lexical Units in FrameNet. After that, images and sentences are loaded into the interface where the human annotation happens.

\begin{figure*}
\centering
\includegraphics[width=\textwidth]{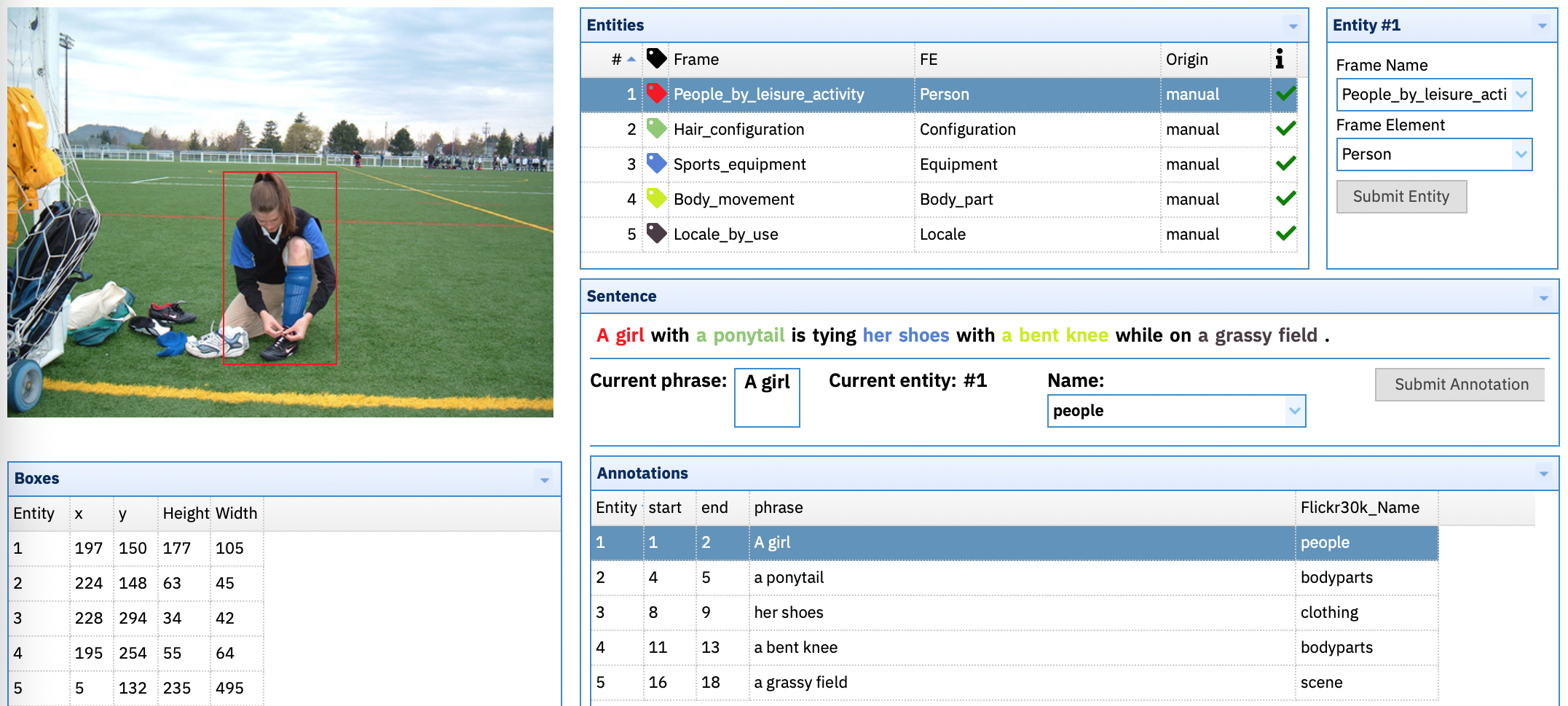}
\caption{\label{fig:caption-anno}User interface for the annotation of Picture-Caption pairings.}
\end{figure*}

\subsubsection{Picture Annotation Process}

Figure \ref{fig:caption-anno} presents the static mode interface, used for the annotation of Picture-Caption pairings. This annotation interface is composed of several panels that are loaded depending on the type of corpus or annotation task being developed. The upper left corner of the interface offers a view of the uploaded image. The panel titled Boxes shows the coordinates for the bounding boxes related to each object/entity being annotated in that picture. The Annotations panel shows the correlations between each object/entity in the image, its co-referenced phrase extracted from the sentence in the middle panel, and the class used to label this object in the original dataset. Finally, the panels Entities and Object are the ones used by the human annotator to assign a Semantic Frame and a FE to each picture-text pair composed by the object/entity in the bounding box and the highlighted phrase in the sentence.

Example sentence (\ref{ex:girl}) has the phrases ``A girl'', ``a ponytail'', ``her shoes'', ``a bent knee'', and ``a grassy field'' correlated with five distinct objects/entities in the image. For the phrase ``A girl'', corresponding to the Entity 1 in the image, the annotator assigned the frame \texttt{People\_by\_leisure\_activity} and the FE \textsc{Person}.

\ex\label{ex:girl}A girl in a ponytail is tying her shoes with a bent knee while on a grassy field.
\xe

This annotation mode generates an XML file, allowing the output to be used with other multimodal annotation tools and integrated with existing transcriptions and annotations from other modules and databases in the FN-Br Webtool environment.

\subsection{Annotation of Videos}

Two types of media are involved in the annotation of videos in FrameNet: audio and image. Therefore, this module of Charon was designed to pre-process videos by (i) extracting verbal data from both audio and images (i. e. subtitles) and deliver it for annotation in the FN-Br Webtool; and (ii) submitting image to an external computer vision system that identifies visual objects and make bounding boxes for those objects available for annotation in Charon. In the following subsections we describe the video annotation pipeline.

\subsubsection{Video Corpus Import and Pre-Processing}

The pipeline designed for corpus import and video pre-processing starts with the selection of the video input, which is imported, pre-processed and separated into two data flows: one for the audio and another for the images.

The next step is the selection of the language of the verbal mode. After the language is selected, the audio data runs through a speech-to-text cloud service, which detects word by word what is said throughout the video.\footnote{For the current implementation, Google Cloud Speech API (\url{https://cloud.google.com/speech-to-text}) is used.} Each word receives time stamps indicating the time span during which they are spoken. 

From the image flow, subtitles are extracted using an optical character recognition software. They are time-stamped and then merged to the text corpus with the output of the speech-to-text software.\footnote{For the current implementation,  Tesseract OCR (\url{https://github.com/tesseract-ocr/tesseract}) is used.} Words and sentences extracted then go through a human-in-the-loop stage, where users can build sentences from the words, edit them, as well as check and adjust time stamps. Finally, the textual part of the corpus is saved and sent to the FN-Br Webtool for annotation.

Charon also processes non-verbal visual data. The images extracted at a 25 frames per second rate are stamped for both time (in seconds) and video frame (in sequential numbers). They run through a computer vision algorithm, which automatically tags objects in each frame, associating a bounding box and a category to them. \footnote{For the current implementation, YOLOv3 \cite{redmon2018yolov3}, trained on the COCO dataset \cite{10.1007/978-3-319-10602-1_48} is used.}

At the end of the pipeline, annotators access the video annotation module, where they visualize both the annotated sentences and the automatically detected objects. This module is described next.

\begin{figure*}[!h]
\centering
\includegraphics[width=\textwidth]{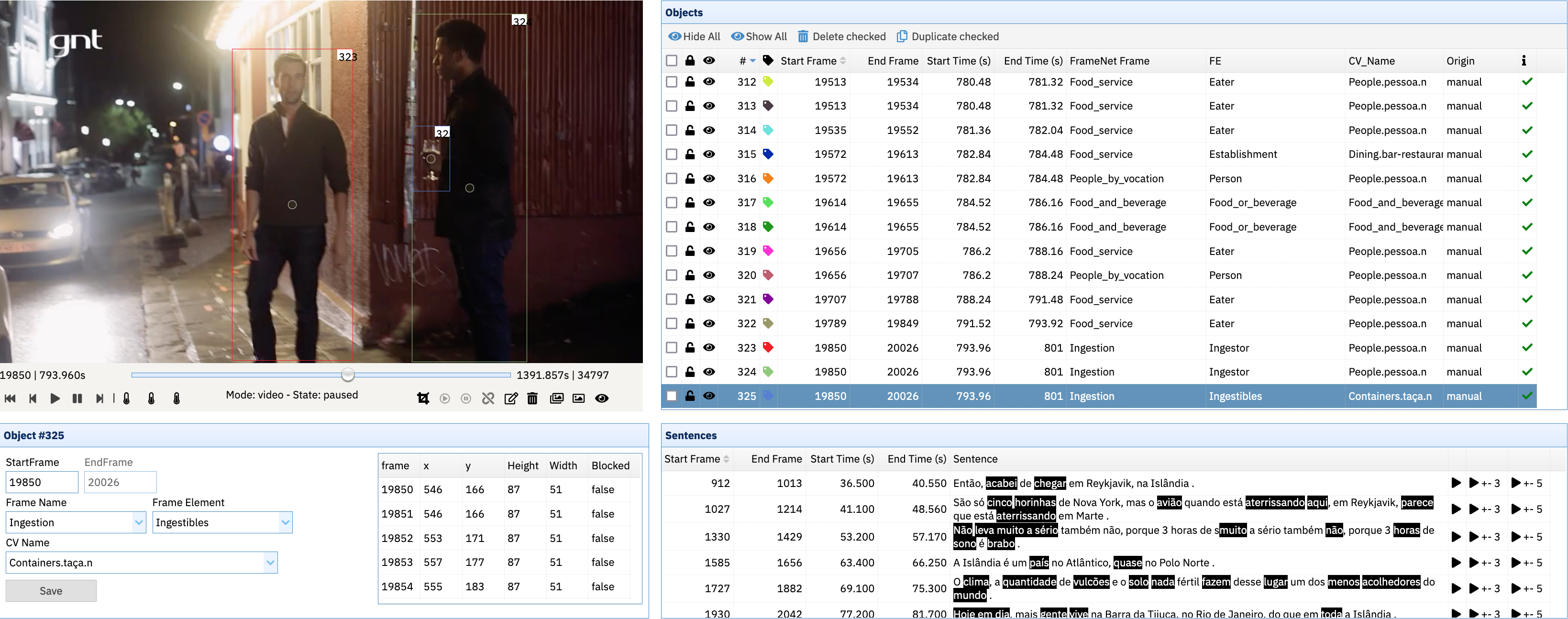}
\caption{\label{fig:Charon_Video_02}Example of video annotation.}
\end{figure*}

\subsubsection{Video Annotation Process}

Charon provides a myriad of possibilities for video annotation by human users, in terms of both methodologies and goals. So far, it has been used to annotate and compare semantic frames evoked by visual objects with those evoked by LUs in sentences. This is why the video annotation module features not only the annotation tools for tagging images, but also the visualization of the sentences annotated in the FN-Br WebTool for the same corpus.

Human annotators can start by reviewing the objects automatically detected by the computer vision software. If annotators agree with the bounding box drawn by the CV software, they select the object in the panel, then use the edit tracking button in the player to link the bounding box to the object through the following video frames. Once the object is not visible anymore or there is a cut point, the annotator presses the pause tracking button, and then the end object one. If annotators do not agree with the bounding box drawn, they can select the object in the panel and delete it.

To create new objects, annotators use the new object button, draw the bounding box over the object they want to detect, then start tracking it. Tracking can be executed manually, frame by frame, or automatically, using the start tracking button. In both cases, annotators determine the end point for the bounding box when the object is not visible anymore or there is a cut point.

Next, annotators have to manually attribute a Semantic Frame and a FE to the object. They choose the frame from the list under the Frame Name field. Once the frame is chosen, a list of its FEs is loaded in the Frame Element field. Annotators should also attribute a Computer Vision name to the object or confirm the label automatically assigned by the computer vision software. This category associates one LU with the object, considering its value as an entity recognizable by computer vision tools or algorithms. In the CV Name field, users may choose from any LU in the framenet database they are using. Figure \ref{fig:Charon_Video_02} shows an example of video annotation. At the moment the image in Figure \ref{fig:Charon_Video_02} is seen on screen, viewers listen one of the men speaking the sentence annotated as in (\ref{ex:speech}):

\ex \label{ex:speech}Bom\textsuperscript{\texttt{Desirability}} que aqui\textsuperscript{\texttt{Locative\_relation}} a gente bebe\textsuperscript{\texttt{Ingestion}} e vai esquentando\textsuperscript{\texttt{Change\_of\_temperature}}, né?\\

\emph{It's good that here we drink and warm ourselves up, innit?}
\xe

\begin{figure}
    \centering
    \includegraphics[width=\columnwidth]{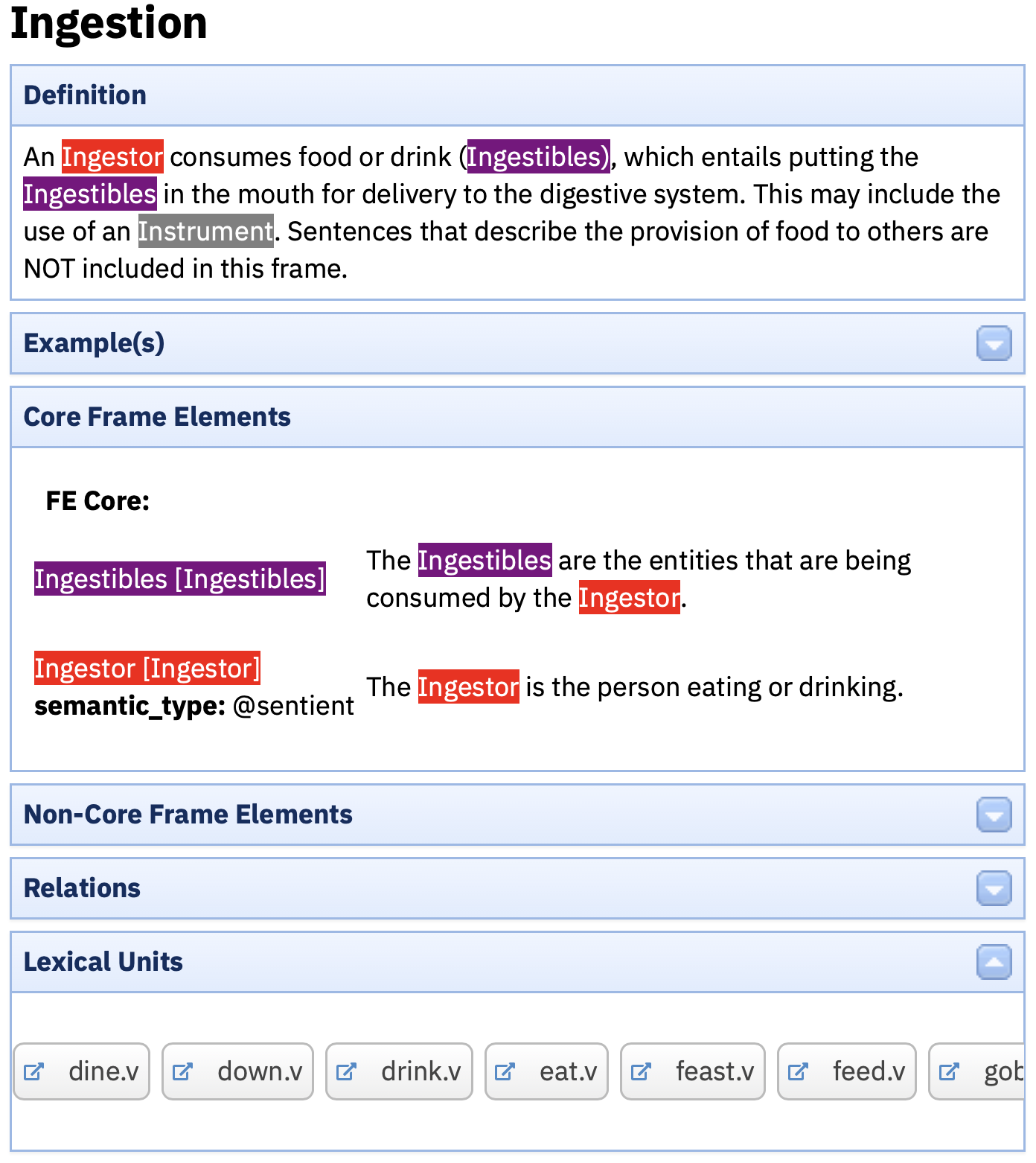}
    \caption{The \texttt{Ingestion} frame.}
    \label{fig:ingestion}
\end{figure}

When looking for correspondences between text and image, objects 323 and 324 were annotated as the \textsc{Ingestors} for the \texttt{Ingestion} frame (Figure \ref{fig:ingestion}). On the other hand, as what is visually recognizable are two human figures, the CV Names chosen were person.n in the \texttt{People} frame. Object 325 was annotated as the \textsc{Ingestibles} in the \texttt{Ingestion} frame and as glass.n in the \texttt{Container} frame for the CV Name. What is interesting here is that in the sentence there is no mention to the \textsc{Ingestibles} FE – it is a null instantiation, – neither to the \texttt{Container} Frame. Therefore, this example shows how meaning layers and granularity can be added to the FrameNet semantic representation by annotating visual data in correspondence with textual data in a corpus.

\section{Expected Datasets}

As demonstrated so far, the addition of other communicative modes to FrameNet annotation allows for building fine-grained semantically annotated multimodal datasets. Two datasets are being currently built by means of Charon's annotation affordances: the Framed Multi 30k and the Frame\textsuperscript{2} datasets \cite{10.3389/fpsyg.2022.838441}.

The Framed Multi 30k Dataset will consist of an improved version of two datasets: the Multi30k dataset \cite{elliott-etal-2016-multi30k} – a multilingual extension of the popular dataset for sentence-based image description Flickr30k \cite{young-etal-2014-image}, – and Flickr30k Entities \cite{7410660}. For each of the 276,000 bounding boxes from Flickr30K Entities, our Framed Multi 30k dataset will add five new sets of Entity-Frame-Frame Element relations, 155,070 new Brazilian Portuguese descriptions, and 155,070 new English-Portuguese translated descriptions.

The Frame\textsuperscript{2} dataset, in turn, is being built to provide means to analyze the interaction between the frame-based semantic representation of verbal language and that produced by the frame-based annotation of video sequences, i.e. sequences of visual frames related with audio, forming a video. The aim is to make it possible to analyze audio and video combination possibilities in terms of frames, as in the example shown in Figure \ref{fig:Charon_Video_02}. This dataset is composed by the multimodal objects selected for annotation in the corpus of the TV Travel Series “Pedro pelo Mundo.” The first data release of Frame\textsuperscript{2} will comprise the annotation of all 10 episodes of the show's first season. This means approximately 12,200 annotation sets for text and 5,000 for image.

\section{Conclusion}

Charon is a unique and robust tool that provides an user-friendly, web-based  interface for fine-grained semantic annotation of both static and dynamic multimodal corpora. The integration with the ever-growing network of semantic frames provided by framenets worldwide allows for large-scale multimodal data analysis. While the current release has already demonstrated its usefulness, many updates and extensions are in the works. A priority is to improve the integration with metadata obtained from machine vision models for automatic object detection.

\section{Acknowledgments}

Authors acknowledge the support of the Graduate Program in Linguistics at the Federal University of Juiz de Fora , as well as the role of Oliver Czulo and Mark Turner in co-supervising, respectively, the doctoral research by Marcelo Viridiano and Frederico Belcavello, whose contributions are reported in this paper. Research presented in this paper was funded by CAPES PROBRAL grant 88887.144043/2017-00 and CNPq grants 408269/2021-9 and 315749/2021-0. Viridiano's research was funded by CAPES PROBRAL PhD exchange grant 88887.628830/2021-00. Belcavello's research was funded by CAPES PDSE PhD exchange grant 88881.362052/2019-01. Authors also acknowledge the contributions of E. P. Hackett\footnote{\url{https://summerofcode.withgoogle.com/archive/2019/projects/5902293138931712}} and Prishita Ray\footnote{\url{https://summerofcode.withgoogle.com/archive/2020/projects/4857286331203584}} to the initial development phase of Charon. Both projects were developed and funded under the Google Summer of Code Program. 


\section{Bibliographical References}\label{reference}

\bibliographystyle{lrec2022-bib}
\bibliography{lrec2022-example}


\end{document}